\documentclass{article}
\usepackage{spconf,amsmath,amssymb,graphicx}
\usepackage{cite,booktabs,multirow}
\usepackage[table]{xcolor}
\usepackage[ruled,vlined]{algorithm2e}
\usepackage{hyperref}

\title{PTC-Bias: Phoneme-Level Temporal Competition for Bias Retrieval and Post-Decoding Correction in Speech LLMs}

\name{
    Zhiqi Ai\textsuperscript{1},
    Han Cheng\textsuperscript{1},
    Shiyi Mu\textsuperscript{1},
    Yongjin Zhou\textsuperscript{1,*},
    Shugong Xu\textsuperscript{2,*}\thanks{\textsuperscript{*}Corresponding authors}
}

\address{
\textsuperscript{1}Shanghai University, Shanghai, China \\
\textsuperscript{2}Xi'an Jiaotong-Liverpool University, Suzhou, China
}

\begin{document}
\ninept
\maketitle
\begin{abstract}
Contextual biasing improves rare-word recognition in speech large language
models (SpeechLLMs), but efficiently exploiting large bias lists remains
challenging. We propose \textbf{PTC-Bias}, a two-stage framework based on
phoneme-level temporal competition. At the prefill stage,
\textbf{PTC Retrieval} performs frame-synchronous phoneme decoding and
temporal competition among candidate pronunciations, producing a compact
bias-word shortlist and corresponding speech intervals. After SpeechLLM
decoding, \textbf{PTC Correction} conducts a second local competition
between the retrieved candidates and mismatched transcript spans within
these intervals. Selective correction reduces near-homophone and
word-segmentation errors while preserving correct transcriptions. Both
stages share the same phoneme posteriors and require no additional SpeechLLM
forward pass. Experiments on LibriSpeech show consistent gains across two
SpeechLLMs and bias lists of up to 2000 words. With Prompt-SLAM-ASR-7B and
2000 bias words, PTC-Bias reduces B-WER by 23.4\%/23.9\% relative to
CTC-Filter on test-clean/test-other, while keeping U-WER nearly unchanged.
\end{abstract}
\section{Introduction}
\label{sec:intro}

Speech large language models (SpeechLLMs) have achieved strong performance
in automatic speech recognition (ASR), but they still struggle with rare
words such as names, locations, and technical terms. Contextual ASR addresses
this problem by incorporating a user-provided bias list through deep biasing,
text-based adaptation, or contextual prompting
~\cite{le2021deepbias,huang2023ustr,qiu2023large,gong2024contextual}.
For SpeechLLMs, however, directly inserting a large bias list into the prompt
increases inference cost and may introduce lexical interference, although
only a small fraction of the list is usually relevant to a given utterance.
This has motivated retrieval-based approaches that select a compact set of
acoustically relevant bias words before SpeechLLM decoding
~\cite{yang2024ctc,lei2025phonetic,gong2025brasr}.

Existing bias-word selection methods rely on phonetic similarity,
speech--text representations, intermediate ASR hypotheses, or learned
retrieval scores
~\cite{yang2024ctc,lei2025phonetic,gong2025brasr,hou2025ranking,
huang2024retrieval,flemotomos2025vq}. Embedding-based methods scale
efficiently to large bias lists but often lack precise temporal localization,
whereas hypothesis-based methods may miss a relevant word if it is absent
from the initial transcription. Moreover, most methods rank candidates
independently, allowing multiple near-homophones associated with the same
speech segment to be selected together rather than compared directly.

Phoneme-based keyword spotting offers finer frame-level evidence. CTC-based
word spotters search candidate pronunciations over phoneme posteriors
~\cite{andrusenko2024ctcws,nakagome2025wctc}, while recent user-defined
keyword-spotting systems employ streaming phoneme search, multi-stage
matching, or frame-asynchronous decoding
~\cite{ai2025dskws,xi2025mfakws,ai2026dmakws}. However, these methods are
primarily designed to detect individual keywords and typically evaluate each
candidate through a separate decoding or matching path, without explicitly
modeling competition between overlapping pronunciations.

Retrieval alone is also insufficient for reliable contextual correction.
Even when the correct bias word is included in the prompt, a SpeechLLM may
still output a frequent near-homophone or produce an incorrect word
segmentation. Existing word-spotting approaches can insert detected keywords
into ASR hypotheses
~\cite{andrusenko2024ctcws,agrawal2025sam}, but a false detection may lead to
an unnecessary replacement of an already correct transcription. Reliable
correction therefore requires a second comparison between the retrieved bias
word and the corresponding transcript span, using acoustic evidence localized
to the detected speech interval.

To address these limitations, we propose \textbf{PTC-Bias}\footnote{Project code: \url{https://github.com/aizhiqi-work/PTC-Bias}}, a two-stage
contextual biasing framework based on phoneme-level temporal competition.
During prefill, \textbf{PTC Retrieval} performs frame-synchronous phoneme
decoding and temporal competition among candidate pronunciations, producing
a compact bias-word shortlist together with their speech intervals. After
SpeechLLM decoding, \textbf{PTC Correction} reuses the detected intervals and
cached phoneme posteriors to conduct a second local competition between each
retrieved candidate and its mismatched transcript span. The two stages thus
use the same phoneme-level acoustic evidence and require no additional
SpeechLLM forward pass. Experiments on LibriSpeech demonstrate consistent
improvements over the corresponding SpeechLLM and CTC-Filter baselines across
two SpeechLLMs and bias-list sizes ranging from 100 to 2000. With
Prompt-SLAM-ASR-7B and 2000 bias words, PTC-Bias reduces B-WER by
23.4\% and 23.9\% relative to CTC-Filter on test-clean and test-other,
respectively, while keeping U-WER nearly unchanged. 

\section{Proposed Method}
\label{sec:method}

\begin{figure*}[t]
    \centering
    \includegraphics[width=\linewidth]{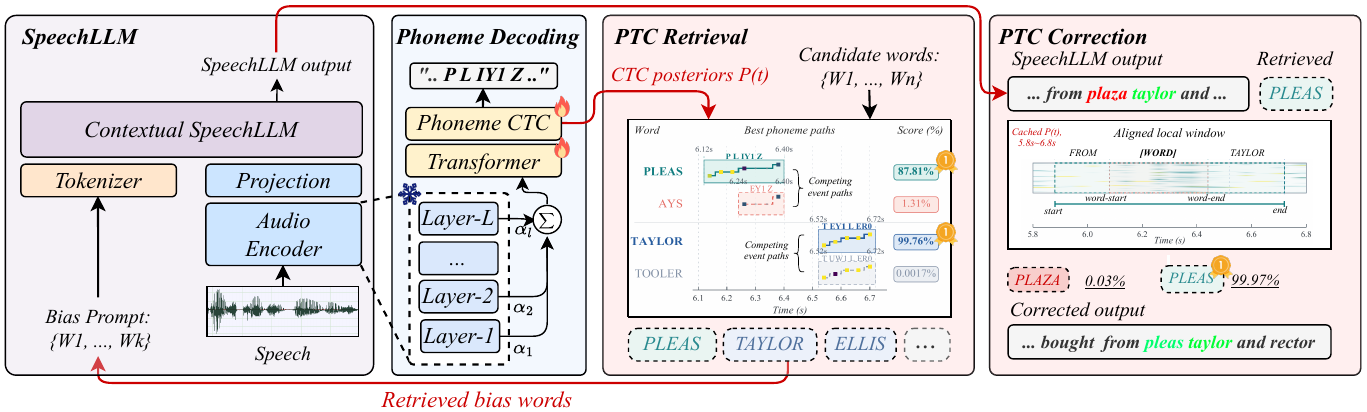}
    \caption{
    Overview of PTC-Bias. A lightweight phoneme-CTC branch extracts
    frame-level phoneme posteriors from the frozen audio encoder.
    PTC Retrieval performs temporal competition to select acoustically
    supported bias words and locate their speech intervals before SpeechLLM
    decoding. PTC Correction then reuses the intervals and cached posteriors
    to correct mismatched transcript spans through local competition.
    }
    \label{fig:overview}
\end{figure*}

\subsection{Overview}
\label{sec:method_overview}

As shown in Fig.~\ref{fig:overview}, PTC-Bias uses a lightweight phoneme-CTC branch to obtain frame-level phoneme posteriors. Before decoding, PTC Retrieval selects bias words and their speech intervals via temporal competition for prompting; after decoding, PTC Correction uses the same posteriors to correct mismatched transcript spans within those intervals. Both stages share the phoneme posteriors and require no additional audio-encoder or SpeechLLM forward pass.

\subsection{Contextual SpeechLLM}
\label{sec:contextual_speechllm}

The Contextual SpeechLLM comprises an audio encoder, an audio projection
module, and an autoregressive LLM decoder. The projected speech
representations and a textual bias prompt are jointly provided to the LLM
for transcription. Since a large bias list increases the prompt length and
may introduce lexical interference~\cite{gong2025brasr}, we include at
most $K$ words selected by PTC Retrieval
(Sec.~\ref{sec:ptc_retrieval}) and retain their speech intervals for
subsequent verification. This preserves the original SpeechLLM interface
without modifying its parameters. Remaining errors are handled by
PTC Correction (Sec.~\ref{sec:ptc_correction}).

\subsection{Phoneme Decoding}
\label{sec:phoneme_decoding}

We attach a lightweight phoneme-CTC branch to the frozen audio encoder.
Its input is a learned weighted sum of all $L$ Transformer-layer outputs:
\begin{equation}
\overline{\mathbf{H}}_t
= \sum_{\ell=1}^{L}\alpha_\ell\mathbf{H}^{(\ell)}_t,
\qquad
\boldsymbol{\alpha}=\operatorname{softmax}(\mathbf{g}),
\label{eq:layer_fusion}
\end{equation}
where $\mathbf{H}^{(\ell)}_t$ is the representation at frame $t$ from
layer $\ell$, and $\alpha_\ell$ is its normalized weight. A lightweight
Transformer and phoneme-CTC head then produce frame-level phoneme
posteriors. Reference transcripts and bias words are converted into
stress-marked ARPAbet sequences using g2pE~\cite{g2pE2019}, providing a
shared phoneme vocabulary for training and retrieval. Only the fusion
weights, lightweight Transformer, and CTC head are optimized. The resulting
posteriors are computed once and shared by PTC Retrieval and PTC Correction.

\begin{table*}[t]
\caption{Contextual ASR results (\%) on LibriSpeech with different
numbers of distractors. Each entry reports
test-clean\,/\,test-other. In the Qwen3-ASR block, $\dagger$
denotes WavLM-based phoneme posteriors; otherwise, the AuT
front-end is used. Lower is better.}
\label{tab:contextual-asr-results}
\centering
\begingroup
\setlength{\tabcolsep}{1.8pt}
\renewcommand{\arraystretch}{1.06}
\newcommand{\cc}[2]{#1\,/\,#2}

\resizebox{\textwidth}{!}{%
\begin{tabular}{@{}lcccccccccccc@{}}
\toprule
\multirow{2}{*}{Contextual ASR Model}
& \multicolumn{3}{c}{$N=100$}
& \multicolumn{3}{c}{$N=500$}
& \multicolumn{3}{c}{$N=1000$}
& \multicolumn{3}{c}{$N=2000$} \\
\cmidrule(lr){2-4}
\cmidrule(lr){5-7}
\cmidrule(lr){8-10}
\cmidrule(lr){11-13}
& WER & B-WER & U-WER
& WER & B-WER & U-WER
& WER & B-WER & U-WER
& WER & B-WER & U-WER \\
\midrule

DB-NNLM~\cite{le2021deepbias}
& \cc{1.98}{5.86} & \cc{5.70}{14.10} & \cc{1.50}{4.90}
& \cc{2.09}{6.09} & \cc{6.20}{15.10} & \cc{1.60}{5.10}
& \cc{2.14}{6.35} & \cc{6.70}{17.20} & \cc{1.60}{5.10}
& \cc{2.27}{6.58} & \cc{7.30}{18.90} & \cc{1.60}{5.20} \\

USTR-CT~\cite{qiu2023large}
& \cc{2.06}{5.38} & \cc{2.00}{4.40} & \cc{2.10}{5.50}
& \cc{2.09}{5.62} & \cc{2.20}{5.60} & \cc{2.10}{5.60}
& \cc{2.16}{5.75} & \cc{2.50}{6.30} & \cc{2.10}{5.70}
& \cc{2.17}{5.84} & \cc{3.00}{7.60} & \cc{2.10}{5.60} \\

CB-QwenAudio~\cite{gong2024contextual}
& \cc{1.60}{3.80} & \cc{5.50}{13.50} & \cc{1.30}{2.60}
& \cc{1.90}{3.90} & \cc{6.00}{14.20} & \cc{1.40}{2.70}
& -- & -- & --
& -- & -- & -- \\

\midrule

Prompt-SLAM-ASR-7B~\cite{yang2024ctc}
& \cc{7.40}{17.90} & \cc{24.80}{44.10} & --
& -- & -- & --
& -- & -- & --
& -- & -- & -- \\

\quad + CTC-Filter~\cite{yang2024ctc}
& \cc{1.27}{2.72} & \cc{3.67}{8.02} & \cc{1.00}{2.16}
& \cc{1.33}{3.04} & \cc{3.92}{9.04} & \cc{1.03}{2.40}
& \cc{1.33}{2.99} & \cc{4.16}{9.33} & \cc{1.00}{2.31}
& \cc{1.38}{3.20} & \cc{4.41}{10.02} & \cc{1.03}{2.47} \\

\quad + Bias Retrieval~\cite{gong2025brasr}
& \cc{1.30}{2.70} & \cc{3.80}{8.30} & --
& \cc{1.30}{3.00} & \cc{4.00}{8.70} & --
& \cc{1.30}{3.10} & \cc{4.10}{9.00} & --
& \cc{1.40}{3.10} & \cc{4.20}{9.30} & -- \\

\quad + PTC Retrieval
& \cc{1.19}{2.48} & \cc{3.30}{6.76} & \cc{0.95}{2.02}
& \cc{1.25}{2.64} & \cc{3.56}{7.08} & \cc{0.99}{2.16}
& \cc{1.33}{2.72} & \cc{3.58}{7.41} & \cc{1.06}{2.22}
& \cc{1.36}{3.00} & \cc{3.85}{8.24} & \cc{1.07}{2.44} \\

\qquad + PTC Correction
& \cc{1.13}{2.41} & \cc{2.76}{6.16} & \cc{0.94}{2.01}
& \cc{1.19}{2.58} & \cc{3.01}{6.51} & \cc{0.98}{2.16}
& \cc{1.26}{2.67} & \cc{3.01}{6.90} & \cc{1.05}{2.22}
& \cc{1.30}{2.93} & \cc{3.38}{7.63} & \cc{1.06}{2.43} \\

\midrule

Qwen3-ASR-0.6B~\cite{qwen3asr2026}
& \cc{1.87}{4.12} & \cc{7.14}{15.04} & \cc{1.26}{2.95}
& \cc{2.12}{4.57} & \cc{8.81}{18.78} & \cc{1.35}{3.06}
& -- & -- & --
& -- & -- & -- \\

\quad + PTC Retrieval
& \cc{1.48}{3.16} & \cc{4.25}{7.80} & \cc{1.16}{2.67}
& \cc{1.51}{3.30} & \cc{4.38}{8.63} & \cc{1.18}{2.73}
& \cc{1.52}{3.32} & \cc{4.54}{8.76} & \cc{1.17}{2.74}
& \cc{1.60}{3.42} & \cc{5.08}{9.24} & \cc{1.20}{2.80} \\

\qquad + PTC Correction
& \cc{1.41}{3.07} & \cc{3.63}{7.00} & \cc{1.15}{2.65}
& \cc{1.44}{3.21} & \cc{3.70}{7.73} & \cc{1.18}{2.72}
& \cc{1.46}{3.26} & \cc{3.94}{8.04} & \cc{1.18}{2.75}
& \cc{1.54}{3.37} & \cc{4.45}{8.61} & \cc{1.21}{2.81} \\

\addlinespace[1pt]
\rowcolor{gray!12}
\quad + CTC-Filter\textsuperscript{$\dagger$}
~\cite{yang2024ctc}
& \cc{1.51}{3.25} & \cc{4.39}{8.35} & \cc{1.18}{2.71}
& \cc{1.53}{3.33} & \cc{4.52}{9.14} & \cc{1.18}{2.71}
& \cc{1.56}{3.40} & \cc{4.79}{9.53} & \cc{1.18}{2.74}
& \cc{1.57}{3.41} & \cc{4.81}{9.45} & \cc{1.19}{2.77} \\

\rowcolor{gray!12}
\quad + PTC Retrieval\textsuperscript{$\dagger$}
& \cc{1.47}{3.13} & \cc{4.23}{7.65} & \cc{1.14}{2.64}
& \cc{1.50}{3.21} & \cc{4.30}{8.14} & \cc{1.18}{2.68}
& \cc{1.51}{3.21} & \cc{4.41}{8.16} & \cc{1.17}{2.68}
& \cc{1.56}{3.26} & \cc{4.88}{8.53} & \cc{1.18}{2.70} \\

\rowcolor{gray!12}
\qquad + PTC Correction\textsuperscript{$\dagger$}
& \cc{1.38}{3.03} & \cc{3.47}{6.84} & \cc{1.14}{2.62}
& \cc{1.41}{3.10} & \cc{3.54}{7.24} & \cc{1.17}{2.66}
& \cc{1.42}{3.11} & \cc{3.67}{7.33} & \cc{1.16}{2.66}
& \cc{1.47}{3.17} & \cc{4.10}{7.76} & \cc{1.17}{2.68} \\

\bottomrule
\end{tabular}%
}
\endgroup
\end{table*}

\begin{algorithm}[t]
\small
\caption{PTC Retrieval}
\label{alg:ptc_retrieval}
\KwIn{Phoneme posteriors $\mathbf{P}_{1:T}$; bias list $\mathcal{B}$}
\KwOut{Retrieved events $\mathcal{R}$}

$\mathcal{W}\leftarrow\textsc{G2P}(\mathcal{B})$\;
$\mathcal{G}\leftarrow\textsc{BuildTrie}(\mathcal{W})$\;
$\mathcal{D}\leftarrow\textsc{InitializeCTC}(\mathcal{G})$\;
$\mathcal{E}\leftarrow\varnothing$\;

\For{$t\leftarrow1$ \KwTo $T$}{
    $\mathcal{D}\leftarrow
    \textsc{TrieCTCStep}(\mathcal{D},\mathbf{P}_t)$\;
    $\mathcal{E}\leftarrow
    \textsc{UpdateBestEvents}(\mathcal{E},\mathcal{D},t,\beta)$\;
}

$\mathcal{E}_M\leftarrow\textsc{TopM}(\mathcal{E},M)$\;
$\{q_k\}\leftarrow
\textsc{IntervalFB}(\mathcal{E}_M,\rho,\tau)$\;
$\mathcal{R}\leftarrow
\textsc{TopKByMarginal}
\bigl(\{(\xi_k,q_k):
\xi_k\in\mathcal{E}_M,\ q_k\geq\eta\},K\bigr)$\;

\Return{$\mathcal{R}$}\;
\end{algorithm}

\subsection{PTC Retrieval}
\label{sec:ptc_retrieval}

PTC Retrieval converts the bias words into stress-marked phoneme
sequences and compiles them into a shared prefix trie. Following
CTC-based word spotting and streaming phoneme search
~\cite{andrusenko2024ctcws,nakagome2025wctc,
xi2025mfakws,ai2026dmakws}, it searches the phoneme posteriors
frame by frame. Each candidate may start at any frame, while
candidates sharing a phoneme prefix reuse the same decoding states.
Algorithm~\ref{alg:ptc_retrieval} summarizes the procedure.

For a bias word $v_k$ with pronunciation $\mathbf{w}^{(k)}$ of
length $U_k$, the search retains its best local CTC path and
corresponding interval $[a_k,b_k]$. The length-calibrated score is
\begin{equation}
S_k =
\max_{\substack{1\leq a\leq b\leq T\\
\mathcal{C}(\boldsymbol{\pi}_{a:b})
=\mathbf{w}^{(k)}}}
\left[
\frac{1}{U_k}
\sum_{t=a}^{b}\log P_t(\pi_t)
+\beta\log U_k
\right],
\label{eq:ptc_score}
\end{equation}
where $\mathcal{C}$ is the CTC collapse operator, $P_t(\pi_t)$ is
the posterior probability of CTC symbol $\pi_t$ at frame $t$, and
$\beta$ controls length calibration. The maximizing path determines
the interval $[a_k,b_k]$.

Each bias word contributes its best event
$\xi_k=(v_k,[a_k,b_k],S_k)$. We retain the $M$ highest-scoring
events as $\mathcal{E}_M$ and treat events with overlapping intervals
as competing explanations of the same acoustic evidence. Let
$\mathfrak{C}_M$ denote all subsets of $\mathcal{E}_M$ whose events
have non-overlapping intervals. The empty subset is included to allow
no bias word to be selected. The weight of a compatible event set
$\mathcal{A}\in\mathfrak{C}_M$ is
\begin{equation}
\psi(\mathcal{A}) =
\exp\left(
\sum_{\xi_j\in\mathcal{A}}
\frac{S_j-\rho}{\tau}
\right),
\label{eq:ptc_set_weight}
\end{equation}
where $\rho$ is the per-event reference score and $\tau$ is the
competition temperature. The competition-adjusted confidence of
event $\xi_k$ is its inclusion marginal
\begin{equation}
q_k =
\frac{
\displaystyle
\sum_{\substack{\mathcal{A}\in\mathfrak{C}_M\\
\xi_k\in\mathcal{A}}}
\psi(\mathcal{A})
}{
\displaystyle
\sum_{\mathcal{A}\in\mathfrak{C}_M}
\psi(\mathcal{A})
}.
\label{eq:ptc_marginal}
\end{equation}
These marginals are computed efficiently using interval
forward--backward. Events satisfying $q_k\geq\eta$ are ranked by
$q_k$, and at most $K$ bias words, together with their intervals
and retrieval confidences, are passed to the Contextual SpeechLLM.

\subsection{PTC Correction}
\label{sec:ptc_correction}

A retrieved bias word may still be transcribed as a near-homophone or
split into multiple words. PTC Correction aligns the initial transcript
$\hat{\mathbf{y}}$ to the cached phoneme posteriors and identifies a
short, phonetically similar span $\hat{\mathbf{y}}_{i:j}$ overlapping
each retrieved interval $[a_k,b_k]$. To compare the retrieved word
$v_k$ with this span under the same phonetic context, we form
$\mathbf{z}_k^{+}
=\mathbf{c}_L\oplus\operatorname{G2P}(v_k)\oplus\mathbf{c}_R$
and
$\mathbf{z}_k^{-}
=\mathbf{c}_L\oplus
\operatorname{G2P}(\hat{\mathbf{y}}_{i:j})\oplus\mathbf{c}_R$,
where $\mathbf{c}_L$ and $\mathbf{c}_R$ are shared context phonemes.

Both sequences are scored within a local window $\Omega_k$ extended
by five posterior frames on each side. Their acoustic margin is
\begin{equation}
\Delta_k =
F(\mathbf{z}_k^{+};\Omega_k)
-
F(\mathbf{z}_k^{-};\Omega_k),
\label{eq:ptc_correction}
\end{equation}
where $F$ denotes the CTC log forward probability. A replacement
requires $\Delta_k$ to exceed a threshold and pass lexical and
boundary checks. Exact homophones, which phoneme-CTC cannot
distinguish, are handled only under restricted rare-word or
segmentation conditions. Conflicting edits are resolved using
$\Delta_k$ and the retrieval marginal $q_k$ from
Eq.~\eqref{eq:ptc_marginal}. All comparisons reuse cached posteriors
without an additional audio-encoder or SpeechLLM forward pass.

\section{Experimental Setup}
\label{sec:experiments}

\subsection{Datasets and Protocol}
\label{sec:data}

We train the phoneme-CTC front-ends on the 460-hour clean subset and
the full 960-hour training set of LibriSpeech
~\cite{panayotov2015librispeech}. Evaluation is conducted on
test-clean and test-other, containing 2,620 and 2,939 utterances,
respectively. Following the Rare5k protocol
~\cite{le2021deepbias,gong2025brasr}, each utterance is paired with
its oracle rare words and
$N\in\{100,500,1000,2000\}$ distractors. PTC Retrieval selects at
most ten words from this list and supplies them as a textual prompt
to Prompt-SLAM-ASR-7B~\cite{yang2024ctc} or
Qwen3-ASR-0.6B~\cite{qwen3asr2026}. Both SpeechLLM backbones are
kept frozen.

\subsection{Models and Implementation}
\label{sec:implementation}

We compare DS-KWS~\cite{ai2025dskws,ai2026dmakws},
WavLM~\cite{chen2022wavlm}, and Qwen3-ASR's audio Transformer
(AuT)~\cite{qwen3asr2026} as phoneme front-ends. DS-KWS is a
3.61M-parameter KWS baseline operating at 25 Hz. For WavLM and AuT,
we freeze the encoders and train only a layer mixture, a two-layer
Transformer, and a 71-class phoneme-CTC head. WavLM operates at
50 Hz with 317.31M total and 1.86M trainable parameters, while AuT
operates at 12.5 Hz with 186.48M total and 1.83M trainable parameters.
Training uses WeNet~\cite{yao2021wenet} with a batch size of 8 on four
NVIDIA RTX 3090 GPUs.

PTC Retrieval uses $M=100$, $K=10$, $\beta=1.1$, $\rho=0$,
$\tau=0.25$, and $\eta=0.05$. PTC Correction extends each local
window by five posterior frames on both sides and applies an
acoustic-margin threshold of 2.0. Both stages share the cached
phoneme posteriors.

\subsection{Evaluation Metrics}
\label{sec:metrics}

We report phoneme error rate (PER), overall word error rate (WER),
biased-word error rate (B-WER), and unbiased-word error rate (U-WER).
Following BR-ASR~\cite{gong2025brasr}, retrieval is evaluated using
$\mathrm{Recall}_{B}@99$, the mean number of candidates required to
reach 99\% oracle-word recall; $\mathrm{Recall}_{B}\#50$, the
oracle-word recall within the top 50; and
$\mathrm{Recall}_{H}\#50$, the near-homophone distractor recall
within the top 50.

\section{Experimental Results}
\label{sec:results}
\subsection{Main Results}
\label{sec:main_results}

Table~\ref{tab:contextual-asr-results} compares PTC-Bias with DB-NNLM~\cite{le2021deepbias}, USTR-CT~\cite{qiu2023large}, CB-QwenAudio~\cite{gong2024contextual}, CTC-Filter~\cite{yang2024ctc}, and Bias Retrieval~\cite{gong2025brasr}. At $N=2000$, PTC-Bias with Prompt-SLAM-ASR-7B achieves WERs of 1.30\%/2.93\% and B-WERs of 3.38\%/7.63\% on test-clean/test-other. Compared with CTC-Filter, it reduces B-WER by 23.4\%/23.9\% while keeping U-WER nearly unchanged. With Qwen3-ASR-0.6B and its AuT phoneme front-end~\cite{qwen3asr2026}, PTC-Bias achieves WERs of 1.54\%/3.37\% and B-WERs of 4.45\%/8.61\%. The consistent gains from $N=100$ to $N=2000$ demonstrate its robustness to large bias lists.

The nested PTC Correction rows further show the benefit of post-decoding competition. At $N=2000$, it reduces B-WER from 3.85\%/8.24\% to 3.38\%/7.63\% with Prompt-SLAM-ASR-7B, and from 5.08\%/9.24\% to 4.45\%/8.61\% with Qwen3-ASR. This indicates that retrieving the correct bias word does not always guarantee correct generation, while local competition can further resolve near-homophone confusions and span-boundary errors. The $\dagger$ rows reuse the WavLM~\cite{chen2022wavlm} retrieval results from Prompt-SLAM-ASR for Qwen3-ASR decoding; under this controlled setting, PTC-Bias reduces B-WER by 14.8\%/17.9\% relative to CTC-Filter.

\subsection{Phoneme Modeling and Bias Retrieval}
\label{sec:phoneme_retrieval}

Table~\ref{tab:phoneme-error-rate} compares the phoneme front-ends,
using DS-KWS~\cite{ai2025dskws,ai2026dmakws} as a lightweight
KWS-oriented baseline. With LS-960 training, WavLM achieves PERs of
1.13\%/2.27\% on test-clean/test-other, compared with
2.06\%/5.13\% for AuT and 4.45\%/11.80\% for DS-KWS. In particular,
the AuT results indicate that existing SpeechLLM audio encoders provide
sufficient phoneme discrimination to support both PTC stages.

Table~\ref{tab:retrieval_results} further compares PTC Retrieval with
BR-ASR~\cite{gong2025brasr}.
Relative to acoustic BR-ASR, PTC Retrieval reduces
$\mathrm{Recall}_{B}@99$ from 42.2 to 16.9 and
$\mathrm{Recall}_{H}\#50$ from 69.3\% to 22.0\%, while maintaining
$\mathrm{Recall}_{B}\#50$ at 99.3\%. PTC Retrieval therefore reaches
high oracle-word recall with a smaller shortlist and substantially
reduces near-homophone interference.




\begin{table}[t]
\caption{Phoneme error rates (PER, \%) of different phoneme front-ends
on LibriSpeech. Lower is better.}
\label{tab:phoneme-error-rate}
\centering
\begingroup
\small
\setlength{\tabcolsep}{3.0pt}
\renewcommand{\arraystretch}{1.05}
\begin{tabular}{@{}lcccc@{}}
\toprule
\multirow{2}{*}{Front-end}
& \multicolumn{2}{c}{LS-460}
& \multicolumn{2}{c}{LS-960} \\
\cmidrule(lr){2-3}
\cmidrule(lr){4-5}
& test-clean & test-other & test-clean & test-other \\
\midrule
DS-KWS~\cite{ai2025dskws,ai2026dmakws}
& 4.44 & 13.39 & 4.45 & 11.80 \\
AuT~\cite{qwen3asr2026}
& 2.39 & 5.88 & 2.06 & 5.13 \\
WavLM~\cite{chen2022wavlm}
& \textbf{1.29} & \textbf{2.44}
& \textbf{1.13} & \textbf{2.27} \\
\bottomrule
\end{tabular}
\endgroup
\end{table}

\begin{table}[t]
\caption{Bias retrieval on LibriSpeech test-other with $N=2000$
distractors. Acoustic and Textual are BR-ASR~\cite{gong2025brasr} variants
.}
\label{tab:retrieval_results}
\centering
\begingroup
\footnotesize
\setlength{\tabcolsep}{1.2pt}
\renewcommand{\arraystretch}{1.08}
\begin{tabular}{@{}lccc@{}}
\toprule
Method
& \shortstack{$\mathrm{Recall}_{B}@99$\\$\downarrow$}
& \shortstack{$\mathrm{Recall}_{B}\#50$ (\%)\\$\uparrow$}
& \shortstack{$\mathrm{Recall}_{H}\#50$ (\%)\\$\downarrow$} \\
\midrule
BR-ASR (Acoustic)~\cite{gong2025brasr}
& 42.2 & \textbf{99.7} & 69.3 \\
BR-ASR (Textual)~\cite{gong2025brasr}
& 47.3 & 99.1 & 58.4 \\
\midrule
PTC Retrieval
& \textbf{16.9} & 99.3 & \textbf{22.0} \\
\bottomrule
\end{tabular}
\endgroup
\end{table}

\subsection{Qualitative and Efficiency Analysis}
\label{sec:efficiency_analysis}

\begin{figure}[!t]
\centering
\includegraphics[width=\linewidth]{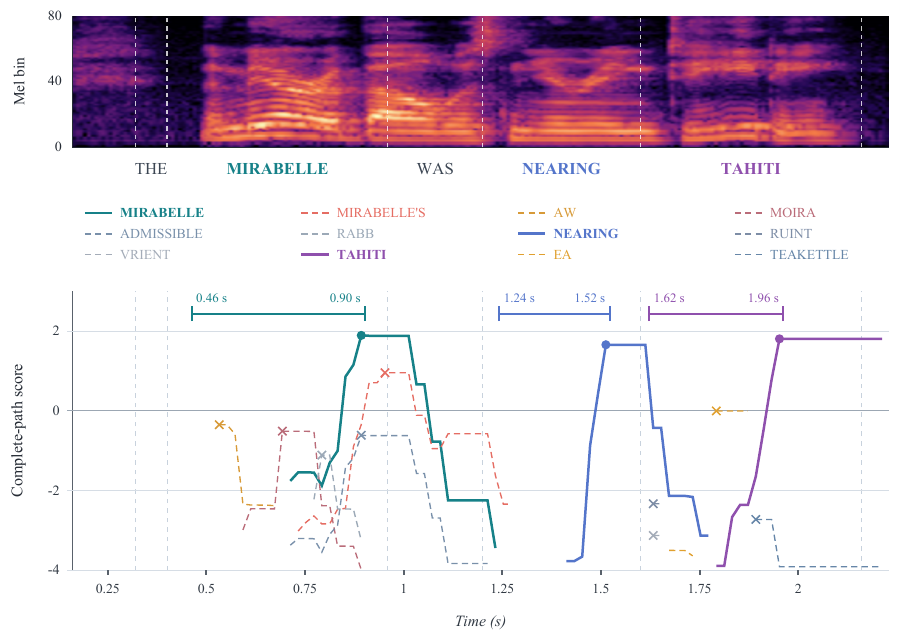}
\caption{Temporal competition among phoneme paths in PTC Retrieval.
Solid paths are selected; dashed paths are competing alternatives.}
\label{fig:ptc-competition}
\end{figure}

Figure~\ref{fig:ptc-competition} visualizes phoneme-level temporal
competition in PTC Retrieval. Similar pronunciations often produce
overlapping high-scoring paths over the same speech segment. By treating
these paths as competing explanations, PTC Retrieval suppresses weaker
near-homophones while retaining confident events at distinct intervals,
producing a compact and less ambiguous shortlist.

\begin{figure}[!t]
\centering
\includegraphics[width=\linewidth]{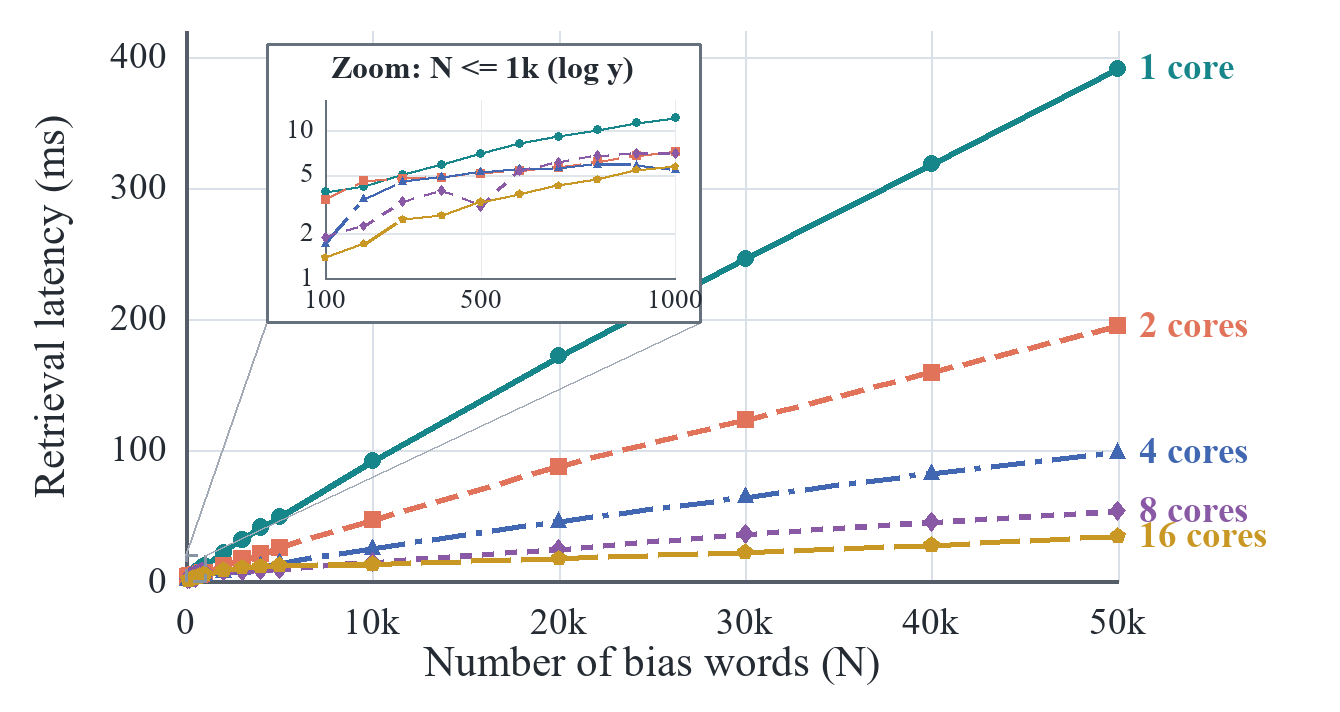}
\caption{CPU search latency of PTC Retrieval for a 5.09-s utterance
across bias-list sizes and worker counts. Phoneme-posterior extraction
and trie construction are excluded.}
\label{fig:ptc_latency}
\end{figure}

Figure~\ref{fig:ptc_latency} reports CPU retrieval latency for a 5.09-s
utterance, excluding phoneme-posterior computation and trie construction.
At $N=2000$, retrieval takes 7.61\,ms with four workers. For
$N=50{,}000$, increasing the worker count from one to 16 reduces latency
from 391.24\,ms to 34.74\,ms. These results show that CPU parallelism
enables PTC Retrieval to scale efficiently to large bias lists.

\section{Conclusion}
\label{sec:conclusion}

We presented PTC-Bias, a two-stage contextual biasing framework for
SpeechLLMs based on phoneme-level temporal competition. Before decoding,
PTC Retrieval uses frame-level phoneme posteriors to select bias words and
locate their speech intervals; after decoding, PTC Correction compares
misrecognized transcript spans with the retrieved candidates. Both stages
reuse the same phoneme posteriors without an additional SpeechLLM forward
pass. Experiments on LibriSpeech show consistent improvements with
100--2000 distractors, different phoneme front-ends, and two SpeechLLM
backbones. PTC-Bias substantially reduces B-WER while largely preserving
U-WER and maintains efficient CPU retrieval for large bias lists.


\bibliographystyle{IEEEbib}
\bibliography{refs}

\end{document}